\pdfoutput=1

\PassOptionsToPackage{table}{xcolor}
\documentclass[11pt]{article}
\usepackage[preprint]{acl}
\usepackage{booktabs}
\usepackage{times}
\usepackage{latexsym}
\usepackage{algpseudocode}
\usepackage[table]{xcolor}
\usepackage{caption}
\usepackage{multirow}
\usepackage{color}
\usepackage{multicol}
\usepackage{array}
\usepackage{arydshln}
\definecolor{Gray}{gray}{0.85}
\definecolor{sent2sent_raw}{RGB}{142,207,201}
\definecolor{sent2sent}{RGB}{211,255,250}
\definecolor{doc2doc_raw}{RGB}{255,204,153}
\definecolor{doc2doc}{RGB}{255,232,205}
\definecolor{refinement_raw}{RGB}{234,235,139}
\definecolor{refinement}{RGB}{217,255,178}
\definecolor{darkred}{RGB}{255,128,128}
\definecolor{lightred}{RGB}{255,217,217}
\usepackage[T1]{fontenc}
\usepackage[utf8]{inputenc}
\usepackage{CJKutf8}
\usepackage{algorithm}
\usepackage{algpseudocode}
\usepackage{microtype}
\usepackage{amsmath}
\usepackage{inconsolata}
\usepackage{amsfonts}
\usepackage{graphicx}
\usepackage{tcolorbox} 

\newcommand{\ignore}[1]{}

\title{Two Intermediate Translations Are Better Than One: Fine-tuning LLMs for Document-level Translation Refinement}

\author{
 \textbf{Yichen Dong\textsuperscript{1}},
 \textbf{Xinglin Lyu\textsuperscript{2}},
 \textbf{Junhui Li\textsuperscript{1}\thanks{Corresponding author: Junhui Li}},
 \textbf{Daimeng Wei\textsuperscript{2}},
\\
 \textbf{Min Zhang\textsuperscript{2}},
 \textbf{Shimin Tao\textsuperscript{2}},
 \textbf{Hao Yang\textsuperscript{2}}
\\
 \textsuperscript{1}School of Computer Science and Technology, Soochow University, Suzhou, China
 \\
 \textsuperscript{2}Huawei Translation Services Center, Beijing, China
\\
\texttt{ycdong@stu.suda.edu.cn,lijunhui@suda.edu.cn}
\\
\texttt{\{lvxinglin1,weidaimeng,zhangmin186,taoshimin,yanghao30\}@huawei.com}
}

\begin{document}
\maketitle
\begin{abstract}
Recent research has shown that large language models (LLMs) can enhance translation quality through self-refinement. In this paper, we build on this idea by extending the refinement from sentence-level to document-level translation, specifically focusing on document-to-document (Doc2Doc) translation refinement. Since sentence-to-sentence (Sent2Sent) and Doc2Doc translation address different aspects of the translation process, we propose fine-tuning LLMs for translation refinement using two intermediate translations, combining the strengths of both Sent2Sent and Doc2Doc. Additionally, recognizing that the quality of intermediate translations varies, we introduce an enhanced fine-tuning method with quality awareness that assigns lower weights to easier translations and higher weights to more difficult ones, enabling the model to focus on challenging translation cases. Experimental results across ten translation tasks with LLaMA-3-8B-Instruct and Mistral-Nemo-Instruct demonstrate the effectiveness of our approach. \footnote{Our code is available at: \url{https://github.com/1078966865/2_better_1/}.}
\end{abstract}

\section{Introduction}

Recent research has shown that large language models (LLMs) can improve their outputs through self-refinement~\cite{madaan-etal-2023-selfrefine}. In machine translation, translation refinement improves translation quality by refining intermediate results. For instance, \citet{chen-etal-2024-iterative} use GPT for translation refinement with simple prompts for iterative improvements. Similarly, \citet{raunak-etal-2023-leveraging} employ a chain of thought (CoT) strategy to describe suggested changes in natural language. \citet{koneru-etal-2024-contextual} futher expand this approach by using document-level context to refine current sentences.

\begin{figure}[!t]
\centering
\includegraphics[width=\columnwidth, trim={0cm 0cm 0cm 0cm}]{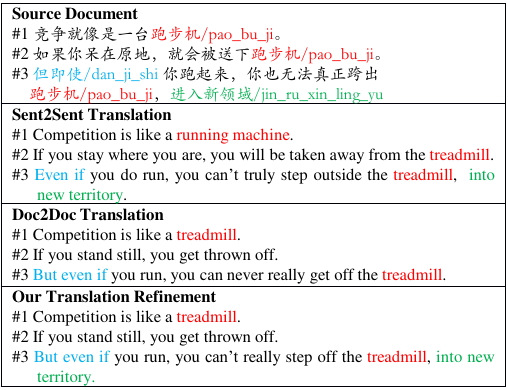}
\caption{An example of Sent2Sent and Doc2Doc Chinese-to-English translations.} 
\label{fig:intro_case}
\end{figure}

\begin{CJK}{UTF8}{gkai}

Unlike previous studies, we extend translation refinement from sentence-level to document-level, refining the translations of all sentences in a document in one go. A document's translation can be generated either by a sentence-to-sentence (Sent2Sent) or document-to-document (Doc2Doc) system. However, Sent2Sent translation, lacking document-level context, often faces discourse-related issues like lexical inconsistency and coherence problems. For example, as shown in Figure~\ref{fig:intro_case}, the word ``跑步机/pao\_bu\_ji'' in the source document is translated as both \textit{running machine} and \textit{treadmill} in the Sent2Sent translation. Additionally, translating ``但即使/dan\_ji\_shi'' as \textit{even if} disrupts coherence by ignoring the discourse relationship between sentences \#2 and \#3. Conversely, while Doc2Doc translation can reduce these discourse-related issues by incorporating both source- and target-side document-level context, it often suffers from under-translation, omitting phrases, clauses, or entire sentences. For example, the verb phrase ``进入新领域/jin\_ru\_xin\_ling\_yu'' in the source document is completely omitted in the Doc2Doc translation.  Taking Chinese-to-English document-level translation as example, Table~\ref{tab:zhen_performance} compares the performance between Sent2Sent and Doc2Doc by LLaMA3-8B-Instruct without fine-tuning. It shows that Doc2Doc achieves better performance in document-level metrics like d-COMET~\cite{vernikos-etal-2022-embarrassingly,rei-etal-2022-comet22}, Coherence~\cite{li-etal-2023-contrastive} and LTCR~\cite{lyu-etal-2021-encouraging}, while Sent2Sent excels in sentence level metrics like ALTI+~\cite{dale-etal-2023-detecting} which detects hallucination and under-translation.\footnote{ Detailed experimental settings, metrics and the results can be found in Section~\ref{sec:experimentation}.}

\begin{table}[t]
\centering
\small
\begin{tabular}{l|ccc|c}
\toprule
\bf System& \bf d-COMET & \bf Coh. & \bf LTCR  & \bf ALTI+ \\
\hline
Sent2Sent& 82.18 & 54.98 & 46.32  & 59.32 \\
\hline
Doc2Doc& 83.60 & 56.21 & 50.00  & 58.66 \\
\bottomrule
\end{tabular}
\caption{Performance comparison between Sent2Sent and Doc2Doc Chinese-to-English translations.}
\label{tab:zhen_performance}
\end{table}

\end{CJK}

Therefore, we conjecture that refining document-level translation over two intermediate translations from both Sent2Sent and Doc2Doc systems can leverage their strengthens, thereby mitigating the aforementioned issues. Given a \textit{source} document, we prompt an existing LLM to generate Sent2Sent and Doc2Doc translations, denoted as \textit{sent2sent} and \textit{doc2doc} translations, respectively. We then construct a document-level refinement quadruple (\textit{source, sent2sent, doc2doc, reference}), where \textit{reference} serves as the naturally refined translation with all the elements at the document level. 

Motivated by \citet{feng-etal-2024-ladder}, who show that distinguishing between sentences with varying quality improves sentence-level translation refinement, we propose an enhanced fine-tuning with quality awareness. This enhanced fine-tuning differentiates instances based on the difficulty of refinement by expanding above quadruple into a quintuple (\textit{source, sent2sent, doc2doc, quality, reference}). The goal of it is to address the varying difficulty of refining translations at sentence- and document-level. Naturally, we weight the documents at sentence level instead of instance level~\cite{lison-bibauw-2017-dialogues} or token level~\cite{fang-feng-2023-cress} since the quality of different sentence within one document may differ significantly. Please refer to Appendix~\ref{apdx:distribution} for more details. By incorporating a quality score as an additional factor during fine-tuning, it helps the model prioritize and output a better translation with differing refinement inputs. 

Overall, our main contributions in this work can be summarized as follows:\footnote{See Appendix \ref{apdx:misalignment} for how our approach can be easily adapted to Doc2Doc translation, even when the source and target documents have differing numbers of sentences.}
\begin{itemize}
\item We extend translation refinement from the traditional sentence-level to the document-level, and further expand it by refining two intermediate translations rather than just one.
\item We introduce enhanced fine-tuning with quality awareness, which differentiates instances based on the difficulty of refinement. 
\item Experimental results on two popular LLMs across ten $X\leftrightarrow$ En document-level translation tasks demonstrate that refining two intermediate translations outperforms refining from a single translation. 
\end{itemize}

\begin{figure*}[!t]
\centering
\resizebox{\textwidth}{!}{
\includegraphics[trim={0cm 0cm 0cm 0cm}]{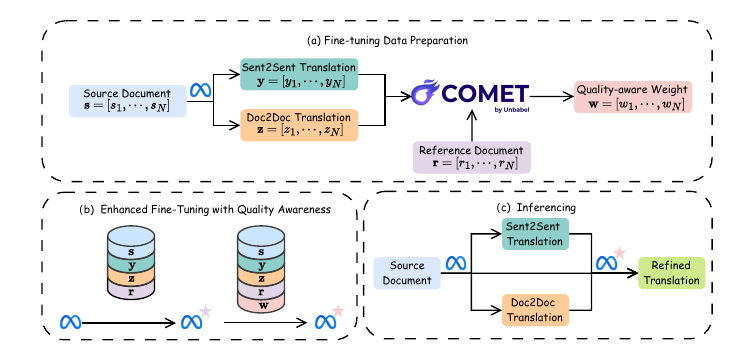}}
\caption{Illustration of our approach. }
\label{fig:main_pipeline}
\end{figure*}

\section{Methodology}
\label{sec:methodology}

Unlike previous studies that fine-tune LLMs for translation using sentence- or document-level parallel datasets, our approach focuses on document-level translation refinement. Specifically, to leverage the  diversity between Sent2Sent and Doc2Doc translations, we introduce document-level translation refinement with two intermediates, using the reference as the target. This emphasis on document-level  refinement, rather than direct translation or sentence-level refinement, distinguishes our work from prior LLM-based translation methods.

As shown in Figure~\ref{fig:main_pipeline}, we develop our document-level refinement LLMs in two steps:
\begin{itemize}
    \item Fine-Tuning Data Preparation (Section~\ref{sec:data_preparation}): For each source-side document in the fine-tuning set, we generate two versions of its translation: one using Sent2Sent translation and the other using Doc2Doc translation.
    \item Enhanced Fine-Tuning with Quality Awareness (Section~\ref{sec:fine_tuning}): Using the prepared fine-tuning data, we fine-tune LLMs in two stages: a na\"ive fine-tuning stage followed by the other stage with a quality-aware strategy.
\end{itemize}
Finally, Section~\ref{sec:inferencing} describes the inference.

\subsection{Fine-Tuning Data Preparation}
\label{sec:data_preparation}

\begin{figure}
    \centering
    \includegraphics[width=1\columnwidth]{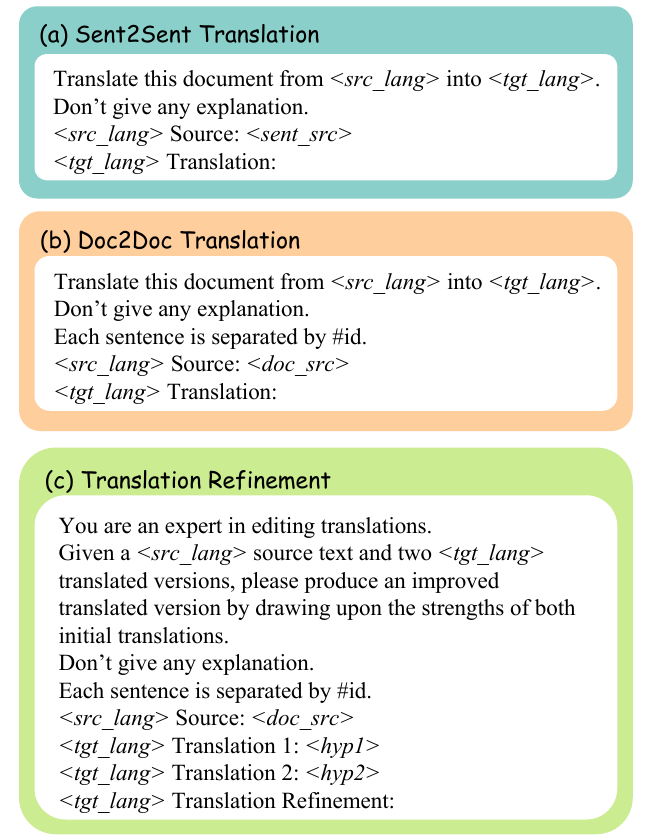}
    \caption{Prompt template used for translation and refinement.}
    \label{fig:prompts}
\end{figure}

We represent a document-level parallel in the fine-tuning data as $\left(\mathbf{s},\mathbf{r}\right)$, where $\mathbf{s}=[s_1, \cdots, s_{N}]$, $\mathbf{r}=[r_1, \cdots, r_{N}]$, with $N$ denoting the number of sentences in the document pair. First, we use LLM $\mathcal{M}_{S}$ to generate sentence-level translations $\mathbf{y}=[y_1, \cdots, y_{N}]$ by translating sentences in $\mathbf{s}$ individually, following the prompt template in Figure~\ref{fig:prompts} (a). Then, we generate document-level translations $\mathbf{z}=[z_1, \cdots, z_{N}]$ by treating the document as a continuous sequence, as shown in  Figure~\ref{fig:prompts} (b). We follow \citet{li-etal-2024-enhancing} to organize the sentences within a document by inserting markers \# id between neighbouring sentences, which indicate their respective positions. Typically, most references $\mathbf{r}$ have higher quality than $\mathbf{y}$ and $\mathbf{z}$ though some references may have lower quality~\cite{xu-etal-2024-contrastive} which can be treated as noise. Thus, we use $\mathbf{r}$ as the target for refinement, as \citet{feng-etal-2024-ladder}. This process yields the document-level refinement quadruple $(\mathbf{s}, \mathbf{y}, \mathbf{z}, \mathbf{r})$. 

\paragraph{Sentence-level Quality-aware Weight.} For two sentences $s_i$ and $s_j$ in document $\mathbf{s}$, the difficulty of refining their translations can vary, depending on the quality of their respective translations $y_i$/$z_i$ and $y_j$/$z_j$. Based on the definition in \citet{feng-etal-2024-ladder}, \textit{easy} translations differ significantly from the reference, providing the most room for refinement,while \textit{hard} translations are nearly perfect, making refinement more challenging. Thus, we assign lower weights to easy translations and higher weights to hard translations. For sentence $s_i$ and its two translations $y_i$ and $z_i$, we use reference-based sentence-level COMET to evaluate the translation quality and compute the weight as follows:
\begin{equation}
\small
\begin{split}
w_i = 1 + \lambda(\max(&\texttt{DA}(s_i, y_i, r_i), \\ 
&\texttt{DA}(s_i, z_i, r_i))-\epsilon ),
\end{split}
\label{eq:weight}
\end{equation}
where $\lambda$ and $\epsilon$ are the hyper-parameters, and \texttt{DA} is computed using reference-based COMET \texttt{wmt22-comet-da}\footnote{ \url{https://huggingface.co/Unbabel/wmt22-comet-da}}~\cite{rei-etal-2022-comet22}. This expands the document-level refinement quadruple into a quintuple $(\mathbf{s}, \mathbf{y}, \mathbf{z}, \mathbf{w}, \mathbf{r})$, where $\mathbf{w}=[w_1,\cdots,w_N]$ represents sentence-level quality-aware weights.\footnote{Comparison with other weighting variants, including instance-level weighting, is provided in Appendix \ref{apdx:weights}.}

 \paragraph{Preventing Position Bias.} 
 Figure~\ref{fig:prompts} (c) shows the prompt template for document-level translation refinement. To avoid position bias, where LLMs might only attend to specific positions~\cite{liu-etal-2024-lost}, the placeholder \textit{<hyp1>} can represent either the sentence-level translation $\mathbf{y}$ or the document-level translation $\mathbf{z}$, with the other in \textit{<hyp2>}. This design creates two instances from the quintuple $(\mathbf{s}, \mathbf{y}, \mathbf{z}, \mathbf{w}, \mathbf{r})$. For clarity, we refer to the quintuple as $(\mathbf{s}, \mathbf{h_1}, \mathbf{h_2}, \mathbf{w}, \mathbf{r})$, where $\mathbf{h_1}$ and $\mathbf{h_2}$ denote the two intermediate translations in the template.

\subsection{Enhanced Fine-Tuning with Quality Awareness}
\label{sec:fine_tuning}
To better leverage the training set, we propose an enhanced fine-tuning strategy, fine-tuning LLM $\mathcal{M_{T}}$ in two stages on the same dataset. In the first stage, we perform na\"ive fine-tuning treating all instances equally. In the second stage, we fine-tune with quality-aware weights. The prompt template for the fine-tuning in both stages is shown in Figure~\ref{fig:prompts} (c). 

\paragraph{Na\"ive Fine-Tuning.} In this stage, the LLM $\mathcal{M_{T}}$ is fine-tuned on the fine-tuning set $\mathcal{T}$ to minimize the following cross-entropy loss function: 

\begin{equation}
\label{eq:loss1}
\small
\begin{split}
&\mathcal{L}_1\left(\mathcal{T}\right)=-\sum_{\mathit{q}\in\mathcal{T}}\log P\left(\mathbf{r} | \mathcal{P}\left(\mathbf{s}, \mathbf{h_1}, \mathbf{h_2}\right)\right)\\
&=-\sum_{\mathit{q}\in\mathcal{T}}\sum_{i=1}^{N}\log P\left(r_i|\mathcal{P}\left(\mathbf{s}, \mathbf{h_1}, \mathbf{h_2}\right), r_{<i}\right),
\end{split}
\end{equation}
where $\mathit{q}$ denotes a quintuple $(\mathbf{s}, \mathbf{h_1}, \mathbf{h_2}, \mathbf{w}, \mathbf{r})$, $\mathcal{P}\left(\mathbf{s}, \mathbf{h_1}, \mathbf{h_2}\right)$ returns the prompt defined by the template, $r_{<i}$ represents the previous sentences before $r_i$ in $\mathbf{r}$. In this stage, all sentences in the reference document $\mathbf{r}$ are assigned equal weights, specifically a weight of 1.

\paragraph{Quality-aware Fine-Tuning.} In this stage, we continue to fine-tune $\mathcal{M}_T$ on $\mathcal{T}$ using a quality-aware strategy, achieved by assigning quality-aware weights to the sentences in the reference $\mathbf{r}$ when calculating the loss function:

\begin{equation}
\label{eq:loss2}
\small
\mathcal{L}_2\left(\mathcal{T}\right)=-\sum_{\mathit{q}\in\mathcal{T}}\sum_{i=1}^{n}w_i\log  P\left(r_i|\mathcal{P}\left(\mathbf{s}, \mathbf{h_1}, \mathbf{h_2}\right), r_{<i}\right).
\end{equation}
Specifically, all tokens within a reference sentence $r_i$ have the same weight $w_i$. And we refer to the fine-tuned LLM as $\mathcal{M}_T^{*}$.

\subsection{Inferencing} 
\label{sec:inferencing}

Once fine-tuning the LLM $\mathcal{M}_T^{*}$ is complete, we use it to refine translations on the test sets. As shown in Figure~\ref{fig:main_pipeline} (c), we first prompt $\mathcal{M}_S$ to generate both Sent2Sent and Doc2Doc translations. Then, for each source document, the two intermediate translations are fed into $\mathcal{M}_T^{*}$ for refinement. During inferencing, quality-aware weights are not needed. 

\section{Experimentation}
\label{sec:experimentation}

\subsection{Experimental Settings}
\paragraph{Datasets.} Following recent works~\cite{li-etal-2024-enhancing,lyu-etal-2024-dempt,alves-etal-2024-tower,cui-etal-2024-efficiently}, to avoid data leakage~\cite{garcia-etal-2023-unreasonable}, we utilize the latest News Commentary v18.1\footnote{ \url{https://www2.statmt.org/wmt24/translation-task.html}} in WMT24, which features parallel text with document boundaries. Our experiments cover five language pairs in both directions: English (En) $\leftrightarrow$ \{German (De), Russian (Ru), Spanish (Es), Chinese (Zh), and French (Fr)\}. For each pair, we randomly select 150 documents for development and another 150 for testing. Specifically, we split long documents into chunks. Details on the dataset and handling long documents are in Appendices~\ref{apdx:dataset} and~\ref{apdx:long_documents}.

\paragraph{Models and Settings.} We select LLaMA-3-8B-Instruct\footnote{ \url{https://huggingface.co/meta-llama/Meta-Llama-3-8B-Instruct}}~\cite{metaai-etal-llama3-2024} and Mistral-Nemo-Instruct\footnote{ \url{https://huggingface.co/mistralai/Mistral-Nemo-Instruct-2407}}~\cite{mistralai-etal-mistral-2024} as the foundation open-source LLMs for applying prompt engineering (i.e., $\mathcal{M}_S$) and quality-aware fine-tuning (i.e., $\mathcal{M}_T$).\footnote{We consider $\mathcal{M}_S$ and $\mathcal{M}_T$ to be the same LLM. For further discussion on cases where $\mathcal{M}_S$ and $\mathcal{M}_T$ differ, please refer to Appendix~\ref{apdx:model_agnostic}.} For detailed fine-tuning and hyper-parameter settings, please refer to Appendix~\ref{apdx:fine_tuning} and ~\ref{apdx:hyper_parameter}.

\paragraph{Baselines.} We compare our approach to several baselines:
\begin{itemize}
\item Sent2Sent: As described in Section~\ref{sec:data_preparation}, we prompt $\mathcal{M}_S$ to generate sentence-level translation. In a contrastive setting, we  first fine-tune $\mathcal{M}_S$ at sentence-level translation and then obtain sentence-level translation, referred as Sent2Sent$_\text{tuned}$.
\item Doc2Doc:  As described in Section~\ref{sec:data_preparation}, we prompt $\mathcal{M}_S$ to generate document-level translation. Similarly, Doc2Doc$_\text{tuned}$ refers to document-level translation from fine-tuned $\mathcal{M}_S$ at document-level translation.
\item SentRefine$_\text{sent}$: It is sentence-level translation refinement by fine-tuning $\mathcal{M}_T$ on Sent2Sent, similar to~\citet{chen-etal-2024-iterative}.
\item DocRefine$_\text{sent}$: It is document-level translation refinement by fine-tuning $\mathcal{M}_T$ on Sent2Sent, similar to~\citet{koneru-etal-2024-contextual}.
\item DocRefine$_\text{doc}$: It is also document-level translation refinement by fine-tuning $\mathcal{M}_T$ on Doc2Doc. 
\end{itemize}

Note that SentRefine$_\text{sent}$, DocRefine$_\text{sent}$ and DocRefine$_\text{doc}$ all use one intermediate translation. Please refer to Figure~\ref{fig:prompts_apdx} in Appendix~\ref{apdx:ape_prompts} for detailed prompts. Differently, our approach uses both Sent2Sent and Doc2Doc as intermediate translations.

\paragraph{Evaluation Metrics.} We report document-level COMET (d-COMET) scores proposed by~\citet{vernikos-etal-2022-embarrassingly}. Specifically, we apply reference-based metric \texttt{wmt22-comet-da}~\cite{rei-etal-2022-comet22}. For other tranditional evaluation metrics, including sentence-level COMET (s-COMET), document-level BLEU (d-BLEU), please refer to Appendix~\ref{apdx:other_results}. 

Besides, we also report several additional metrics. 1) We follow \citet{li-etal-2023-contrastive} and \citet{su-etal-2022-contrastive} to compute coherence score (Coh.) using cosine similarity between the sentence embeddings of SimCSE~\cite{gao-etal-2021-simcse} of the neighbouring sentences. 2) We report ALTI+ score~\cite{ferrando-etal-2022-towards,dale-etal-2023-detecting,wu-etal-2024-word} to detect under-translation and hallucination issues in translation.  3) We follow \citet{lyu-etal-2021-encouraging} and compute LTCR score to measure lexical translation consistency. 4) We compute document-level perplexity (PPL) using GPT-2\footnote{\url{https://huggingface.co/openai-community/gpt2}}~\cite{radford-etal-2019-gpt2}.   5) We report BlonDe~\cite{jiang-etal-2022-blonde}, which evaluates discourse phenomena via a set of automatically extracted features~\cite{deutsch-etal-2023-training}. Except for ALTI+, these metrics are document-level metrics. LTCR, BlonDe, and PPL are computed only for the $X \rightarrow$ En translation direction, while the other two metrics are applicable to all translation directions.

\begin{table*}[h]
\centering
\small
\begin{tabular}{l|l|ccccc|ccccc|c}
\toprule
\multirow{2}{*}{\bf \#} &\multirow{2}{*}{\bf System} & \multicolumn{5}{c|}{\bf \textit{X}$\rightarrow$En} & \multicolumn{5}{c|}{\bf En$\rightarrow$\textit{X}} & \multirow{2}{*}{\bf Avg.} \\
& & \bf De$\rightarrow$ & \bf Es$\rightarrow$ & \bf Ru$\rightarrow$ & \bf Fr$\rightarrow$ & \bf Zh$\rightarrow$ & \bf $\rightarrow$ De & \bf $\rightarrow$ Es & \bf $\rightarrow$ Ru & \bf $\rightarrow$ Fr & \bf $\rightarrow$ Zh\\
\hline
\multicolumn{12}{c}{LLaMA-3-8B-Instruct} \\ 
\hline
\cellcolor{sent2sent}1 & \cellcolor{sent2sent}{Sent2Sent} & 85.97 & 86.62 & 81.63 & 84.43 & 82.18 & 82.50 & 85.02 & 80.97 & 82.89 &76.80& 82.90 \\
\cellcolor{sent2sent}2 & \cellcolor{sent2sent}{Sent2Sent$_\text{tuned}$} & 87.94 & 87.46 & 81.98 & 86.46 & 84.18 & 85.42 & 86.11 & 80.88& 84.30 & 82.84 & 84.76\\ 
\cellcolor{doc2doc}3 & \cellcolor{doc2doc}{Doc2Doc} & 87.05 & 87.21 & 81.07 & 85.40 & 83.60 & 83.35 & 85.36 & 80.18 & 83.14 &81.89& 83.83 \\
\cellcolor{doc2doc}4 & \cellcolor{doc2doc}{Doc2Doc$_\text{tuned}$} & 87.82 & 88.04 & 81.25 & 86.37 & 84.88 & 85.45 & 85.61 & 81.06 & 84.63 & 82.18 & 84.73 \\
\hdashline
\cellcolor{sent2sent}5 & \cellcolor{sent2sent}{SentRefine$_\text{sent}$}  &   83.70 & 87.99  &  82.64  &  85.98  &  84.08  &85.21& 86.34 & \underline{83.74}   &84.57 & 82.93 & 84.72 \\
\cellcolor{doc2doc}6 & \cellcolor{doc2doc}{DocRefine$_\text{sent}$}      & 87.42 & 87.98 & 81.16 & \underline{86.56} & 85.06 & 85.38 & 86.32 & 80.39 & 84.43 &82.61& 84.73 \\
\cellcolor{doc2doc}7 & \cellcolor{doc2doc}{DocRefine$_\text{doc}$}       & 87.71 & 88.06 & \underline{82.73} & 86.32 & 84.99 & 85.07 & 86.49 & 83.16 & \underline{84.73} & 82.70 & 85.19 \\ 
\hdashline
\cellcolor{refinement}8 & \cellcolor{refinement}{\bf Ours} & \cellcolor{lightred}\bf 88.14 & \cellcolor{lightred}\bf \cellcolor{lightred}88.42 & \cellcolor{lightred}\bf 82.75 & \cellcolor{lightred}\bf 86.69 &\cellcolor{darkred}\bf 85.39 & \cellcolor{darkred}\bf 86.05 & \cellcolor{lightred}\bf 86.86 & \cellcolor{lightred}\bf 83.85 & \cellcolor{lightred}\bf 84.84 &\cellcolor{darkred}\bf 83.35 & \cellcolor{darkred}\bf 85.63 \\ 
\cellcolor{refinement}9 & \cellcolor{refinement}{ ~~ - \footnotesize{QA } } & \cellcolor{lightred}\underline{88.02} & \cellcolor{lightred}\underline{88.35} & 82.63 & 86.53 & \cellcolor{lightred}\underline{85.09} & \cellcolor{lightred}\underline{85.70} & \cellcolor{lightred}\underline{86.60} & 83.17 & 84.48 & \cellcolor{lightred}\underline{82.98} & \cellcolor{lightred}\underline{85.36} \\
\hline
\multicolumn{12}{c}{Mistral-Nemo-Instruct} \\ \hline
\cellcolor{sent2sent}1& \cellcolor{sent2sent}Sent2Sent & 86.85 & 87.21 & 82.86 & 85.27 &83.82 & 84.66 & 85.47 & 83.78 & 83.67 &79.39 & 84.30 \\
\cellcolor{sent2sent}2& \cellcolor{sent2sent}Sent2Sent$_\text{tuned}$& 86.86& 86.89& 83.33& 85.79 & 83.96 & 85.49 & 85.77 & 84.58 & 84.49 & 81.18 & 84.83 \\ 
\cellcolor{doc2doc}3& \cellcolor{doc2doc}Doc2Doc & 87.61 & 87.64 & 82.60 & 85.95 &84.55& 84.34 & 85.14 & 84.34 & 83.66 &81.34 & 84.72 \\
\cellcolor{doc2doc}4& \cellcolor{doc2doc}Doc2Doc$_\text{tuned}$ & 87.80 & 88.34 & 82.60 & 86.39 & 85.16 &86.50 &86.72 &85.68 & 85.28 & 81.27 & 85.57\\
\hdashline
\cellcolor{sent2sent}5& \cellcolor{sent2sent}SentRefine$_\text{sent}$ & 87.73&88.23& 83.87& 86.23& 84.71 & 86.36& 86.48&\underline{85.63}&85.06&81.27 &85.56 \\
\cellcolor{doc2doc}6& \cellcolor{doc2doc}DocRefine$_\text{sent}$ & 88.09 & \underline{88.50} & 82.34 & 86.21 &\underline{85.40}& 86.58 & 86.91 & 84.67 & 85.09&\underline{84.06}& 85.79 \\
\cellcolor{doc2doc}7& \cellcolor{doc2doc}DocRefine$_\text{doc}$      & \underline{88.13} & 88.37 & 81.65 & \underline{86.41} & 85.20 & 86.44 & \underline{86.95} & 83.90 & 85.11&83.86& 85.61 \\ 
\hdashline
\cellcolor{refinement}8& \cellcolor{refinement}\bf Ours & \cellcolor{lightred}\textbf{88.45} & \cellcolor{darkred}\textbf{88.99} & \cellcolor{darkred}\textbf{84.59} & \cellcolor{darkred}\textbf{87.00} & \cellcolor{darkred}\textbf{85.83} & \cellcolor{lightred}\textbf{86.89} & \cellcolor{lightred}\textbf{87.31} & \cellcolor{lightred}\textbf{85.99} & \cellcolor{lightred}\textbf{85.50} & \cellcolor{darkred}\textbf{84.53} & \cellcolor{darkred}\textbf{86.51} \\
\cellcolor{refinement}9& \cellcolor{refinement}~~ - \footnotesize{QA } & 88.01 & 88.27 & \cellcolor{lightred}\underline{83.89} & 86.40 & 85.37 & \cellcolor{lightred}\underline{86.70} & 86.94& 85.34 & \cellcolor{lightred}\underline{85.43} & 83.86 & \cellcolor{lightred}\underline{86.02} \\ 
\hline
\end{tabular}
\caption{Performance in document-level COMET (d-COMET) score. Bold scores represent the highest performance, while underlined scores indicate the second-best performance. \textit{-QA} indicates disabling the quality-aware fine-tuning stage. Scores of our approach (System \#8 and \#9) that exceed the highest value in the baselines (System \#1 $\sim$ \#7) by $\ge$ 0.4 points are highlighted with \colorbox{darkred}{dark red boxes}, while those that are positive but $<$ 0.4 points higher are highlighted with \colorbox{lightred}{shallow red boxes}. }
\label{tab:main_results}
\end{table*}

\subsection{Main Results}
\label{sub:results}

Table \ref{tab:main_results} presents the performance comparison in d-COMET. From it, we observe:

\begin{itemize}
    \item Extending the translation unit from sentence-level to document-level improves overall performance, as Doc2Doc outperforms Sent2Sent. This aligns with findings from related studies~\cite{karpinska_iyyer_wmt_2023_large}. However, fine-tuned LLMs exhibit different performance trends. LLaMA-3-8B-Instruct shows similar performance for both Sent2Sent$_{\text{tuned}}$ and Doc2Doc$_{\text{tuned}}$, while Mistral-Nemo-Instruct performs better with Doc2Doc$_{\text{tuned}}$ compared to Sent2Sent$_{\text{tuned}}$.
    
    \item Refining with a single input, whether from Sent2Sent or Doc2Doc, leads to higher COMET scores. However, this refinement shows little to no improvement over the performance of directly fine-tuned LLMs. 
    
    \item Our refinement approach, based on the two intermediate translations Sent2Sent and Doc2Doc, significantly improves translation performance across all language pairs. It achieves COMET score improvements of 2.73 and 1.80 on LLaMA-3-8B-Instruct, and 2.21 and 1.79 on Mistral-Nemo-Instruct. Our approach also outperforms other baselines, including both refining with single translations and directly fine-tuning, demonstrating the effectiveness of our proposed approach.
    \item Lastly, disabling the quality-aware fine-tuning stage results in a performance drop, highlighting the effectiveness of our fine-tuning strategy. Additionally, compared to SentRefine$_\text{sent}$, DocRefine$_\text{sent}$, and DocRefine$_\text{doc}$, refinement using two intermediate translations outperforms refinements with just one.
\end{itemize}

\begin{table*}[]
\centering
\small
\begin{tabular}{l|l|ccccc}
\toprule
\bf \# & \textbf{System} & \textbf{Coh.} $\uparrow$  & \textbf{ALTI+} $\uparrow$& \textbf{LTCR} $\uparrow$  & \textbf{PPL} $\downarrow$ & \textbf{BlonDe} $\uparrow$ \\
\hline
\cellcolor{sent2sent}1 & \cellcolor{sent2sent}Sent2Sent  & 56.17  & 42.57& 57.23& 32.86 & 48.49 \\
\cellcolor{sent2sent}2 & \cellcolor{sent2sent} Sent2Sent$_\text{tuned}$  & 56.23  & 42.94& 60.45 & 30.34 & 58.61 \\
\cellcolor{doc2doc}3 &\cellcolor{doc2doc}Doc2Doc  & 62.28  & 40.04 &  61.25 & 31.85 & 51.30 \\
\cellcolor{doc2doc}4 &\cellcolor{doc2doc}Doc2Doc$_\text{tuned}$  & 63.42  & 42.99 & 64.99 & 31.58 & 57.86 \\
\hdashline
\cellcolor{sent2sent}5 &\cellcolor{sent2sent}SentRefine$_\text{sent}$ & 64.27  & \underline{43.09}& 60.08& 32.14 & 57.47 \\
\cellcolor{doc2doc}6 &\cellcolor{doc2doc}DocRefine$_\text{sent}$ & 64.95 & 43.00 &  63.62 & \underline{30.13} & 58.69 \\
\cellcolor{doc2doc}7 &\cellcolor{doc2doc}DocRefine$_\text{doc}$   & 65.09  & 42.80 &  63.68 & 31.62 & 59.01 \\
\hdashline
\cellcolor{refinement}8 &\cellcolor{refinement}Ours  & \textbf{67.12}  & \textbf{43.53}& \textbf{66.57}& \textbf{26.51} & \textbf{59.86} \\ 
\cellcolor{refinement}9 &\cellcolor{refinement} ~~ - \footnotesize{QA} & \underline{66.07}  & 43.06 & \underline{65.98} & 31.64 & \underline{59.57}\\
\bottomrule
\end{tabular}
\caption{\label{tab:discourse_metrics}Averaged performance of LLaMA-3-8B-Instruct in additional metrics. }
\end{table*}

Table~\ref{tab:discourse_metrics} presents the performance on several additional metrics when LLaMA-3-8B-Instruct is used. The results show that, except for ALTI+, document-level translation and refinement systems outperform their sentence-level counterparts. By combining the strengths of Sent2Sent and Doc2Doc translations, our approach achieves the best performance across all five metrics.

\section{Discussion}

\subsection{Refining Translations by GPT and NLLB} 
\label{sec:gpt_nllb}

\begin{table*}[]
\centering
\small
\resizebox{\textwidth}{!}{
\begin{tabular}{l|l|ccccc|ccccc|c}
\toprule
\bf \multirow{2}{*}{\#} & \multirow{2}{*}{\bf System} & \multicolumn{5}{c|}{\bf \textit{X}$\rightarrow$En} & \multicolumn{5}{c|}{\bf En$\rightarrow$\textit{X}} & \multirow{2}{*}{\bf Avg.} \\
 & & \bf De$\rightarrow$ & \bf Es$\rightarrow$ & \bf Ru$\rightarrow$ & \bf Fr$\rightarrow$ & \bf Zh$\rightarrow$ & \bf $\rightarrow$ De & \bf $\rightarrow$ Es & \bf $\rightarrow$ Ru & \bf $\rightarrow$ Fr & \bf $\rightarrow$ Zh\\
\hline
\multicolumn{13}{c}{GPT Translation \& Refining GPT Translation}\\
\hline
\cellcolor{sent2sent}1 &\cellcolor{sent2sent}GPT Sent2Sent  & 86.49 & 86.53 & 82.43 & 84.73 & 83.98 & 85.96 & 86.52 & 85.28 & 84.97 & 83.70 & 85.06 \\
\cellcolor{doc2doc}2 &\cellcolor{doc2doc}GPT Doc2Doc & 87.00 & 87.12 & \underline{83.71} & 85.64 & 84.75 & 86.30 & 86.76 & 85.59 & 85.23 & 84.07 & 85.62 \\
\hdashline
\cellcolor{sent2sent}3 &\cellcolor{sent2sent}GPT SentRefine$_\text{sent}$ & 86.86 & 86.89 & 83.37 & 83.70 &  83.33 & 85.32 & 86.43 & 85.42 & 84.30& 83.99 & 84.96 \\
\cellcolor{doc2doc}4 &\cellcolor{doc2doc}GPT DocRefine$_\text{sent}$ & 87.03 & 87.26 & 83.23 & 85.77 & 84.29 & 86.57 & 87.04 & 86.04 & 85.40 & 84.07 & 85.67\\
\cellcolor{doc2doc}5 &\cellcolor{doc2doc}GPT DocRefine$_\text{doc}$ & 87.04 & 87.29 & 83.27 & 85.63 & 84.41 & 86.37 & 87.03 & 86.14 & 85.43 & 83.93 & 85.62\\ 
\cellcolor{refinement}6 &\cellcolor{refinement}GPT DocRefine$_\text{doc+sent}$ & \cellcolor{lightred}87.39 & \cellcolor{lightred}87.65 & \cellcolor{lightred}83.44 & 85.77 & \cellcolor{lightred}84.78 & \cellcolor{lightred}\underline{86.61} & \underline{86.96} & \cellcolor{lightred}\underline{86.16} & \cellcolor{lightred}\underline{85.46} & \bf \cellcolor{lightred}84.13 & \cellcolor{lightred}\underline{85.84} \\
\hdashline
\cellcolor{refinement}7 &\cellcolor{refinement}L-DocRefine$_\text{doc+sent}$ &\cellcolor{darkred}\underline{87.88} &\cellcolor{darkred}\underline{88.15} &82.07 & \cellcolor{darkred}\underline{86.57} &\cellcolor{darkred}\underline{85.22} & 86.31 &86.09 & 83.66 & 85.28 & 83.32 & 85.46\\
\cellcolor{refinement}8 &\cellcolor{refinement}M-DocRefine$_\text{doc+sent}$ & \cellcolor{darkred}\bf 88.14 & \cellcolor{darkred}\bf 88.22 & \cellcolor{darkred}\bf 84.39 & \cellcolor{darkred}\bf 86.73 & \cellcolor{darkred}\bf 85.48 & \cellcolor{lightred}\bf 86.88 & \cellcolor{lightred}\bf 87.20 & \cellcolor{lightred}\bf 86.20 & \cellcolor{lightred}\bf 85.69 & \underline{84.12} & \cellcolor{darkred}\bf 86.31 \\
\hline
\multicolumn{13}{c}{NLLB Translation \& Refining NLLB Translation}\\
\hline
\cellcolor{sent2sent}9&\cellcolor{sent2sent}NLLB Sent2Sent&86.79& 87.55& \underline{83.22}& 85.62& 83.17& 84.93& 86.22& \underline{85.47} & 84.60& \bf 84.17& 85.17\\
\hdashline
\cellcolor{refinement}10&\cellcolor{refinement}L-DocRefine$_\text{doc+sent}$&\cellcolor{darkred}\underline{87.85}&\cellcolor{darkred}\underline{88.41}&81.65&\cellcolor{darkred}\bf 86.51 & \cellcolor{darkred}\underline{85.00}&\cellcolor{darkred}\underline{86.20}&\cellcolor{lightred}\underline{86.43}&83.03&\cellcolor{lightred}\underline{84.97}&82.80&\cellcolor{lightred}85.29\\
\cellcolor{refinement}11&\cellcolor{refinement}M-DocRefine$_\text{doc+sent}$&\cellcolor{darkred}\bf 88.10&\cellcolor{darkred}\bf 88.66&\cellcolor{darkred}\bf 84.44&\cellcolor{darkred}\underline{86.17}& \cellcolor{darkred}\bf 85.48& \cellcolor{darkred}\bf 86.81& \cellcolor{darkred}\bf 86.74&\cellcolor{lightred}\bf 85.49& \cellcolor{darkred}\bf 85.34& \underline{84.08}& \cellcolor{darkred}\underline{86.13}\\
\bottomrule
\end{tabular}}
\caption{\label{tab:refine_gpt}Performance in d-COMET when refining translations from GPT-4o-mini (upper) and NLLB (lower). For the GPT-based refinement systems, we use the same prompt templates as those used in our approach, but without fine-tuning (System \#3$\sim$\#6). L-* and M-* denote our fine-tuned LLaMA-3-8B-Instrcut and Mistral-Nemo-Instruct (i.e., \textit{Ours} in Table~\ref{tab:main_results}), respectively.}
\end{table*}

To further evaluate our approach, we use our fine-tuned LLMs (i.e., \textit{Ours} in Table~\ref{tab:main_results}) to refine translations from GPT-4o-mini~\cite{openai-etal-gpt4o-2024} and NLLB~\cite{nllb-team-2022-nllb}, the state-of-the-art NMT system. As shown in the upper part of Table~\ref{tab:refine_gpt}, refining GPT-4o-mini’s output with one intermediate translation yields limited improvement ( \#4/\#5 vs. \#2). In contrast, using two intermediate translations increase the COMET score by 0.22 (\#6 vs. \#2), suggesting that using two intermediate translations is more effective. Our two fine-tuned LLMs behave differently: LLaMA-3-8B-Instruct experiences a slight drop (85.62 to 85.46), while Mistral-Nemo-Instruct successfully improves performance (85.62 to 86.31). For detailed s-COMET scores, please refer to Appendix~\ref{apdx:other_results}.

Additionally, we refine NLLB-generated translations using our fine-tuned LLMs.  Since NLLB does not support Doc2Doc translation, we simplify the process by treating Doc2Doc translation as equivalent to Sent2Sent translation (i.e., the two intermediate translations are identical) for refinement with our fine-tuned LLMs. As shown in the lower part of Table~\ref{tab:refine_gpt}, even though the two intermediate translations are identical, LLaMA-3-8B-Instruct still shows a slight improvement (85.17 to 85.29), while Mistral-Nemo-Instruct demonstrates a more substantial improvement, (85.17 to 86.13).

Furthermore, we prompt LLMs to refine translations produced by other LLMs. For detailed experimental results, please refer to Appendix~\ref{apdx:model_agnostic}.

\subsection{Effect of Enhanced Fine-tuning with Quality Awareness}
Table~\ref{tab:two_stage_fine_tuning} compares the performance on En$\leftrightarrow$De and En$\leftrightarrow$Zh directions for various fine-tuning strategies. Removing either the na\"ive or the quality-aware fine-tuning stage reduces performance. Meanwhile, replacing the quality-aware fine-tuning stage with na\"ive one may cause a performance drop, indicating that each stage in our enhanced fine-tuning with quality awareness contributes to the overall performance, which can effectively alleviate overfitting.

\begin{table}[]
\centering
\small
\setlength{\tabcolsep}{4pt}
\begin{tabular}{lccccc}
\toprule
\bf Stage1 & \bf Stage2 & \bf De$\rightarrow$En & \bf En$\rightarrow$De &\bf Zh$\rightarrow$En  &\bf  En$\rightarrow$Zh \\
\hline
na\"ive & QA & \bf 88.14 &  \bf 86.05 & \bf 85.39 &\bf 83.35\\
\hline
na\"ive & - & 88.02 & 85.70 & 85.09 & 82.98\\
QA & - & 87.76 & 85.60 & 84.88 & 83.05 \\
na\"ive & na\"ive & 87.75 & 85.91 & 83.98 & 82.14 \\
\bottomrule
\end{tabular}
\caption{Performance comparison when using different fine-tuning strategies. QA indicates quality-aware fine-tuning.}
\label{tab:two_stage_fine_tuning}
\end{table}

\subsection{Effect of Preventing Position Bias}
To prevent introducing position bias, \textit{<hyp1>} in the prompt template can be either Sent2Sent or Doc2Doc translation. To examine its effect, we compare it with a version where \textit{<hyp1>} is always set to Sent2Sent and \textit{<hyp2>} is set to Doc2Doc. As shown in Table~\ref{tab:position_bias}, preventing position bias leads to a significant boost in performance.

\begin{table}[]
\centering
\small
\begin{tabular}{l|cc}
\toprule
\bf Our Approach  & \bf De$\rightarrow$En & \bf En$\rightarrow$De  \\
\hline
w/ preventing position bias & \bf 88.14 &  \bf 86.05 \\
\hline
w/o preventing position bias & 87.60 & 85.55 \\
\bottomrule
\end{tabular}
\caption{Performance comparison with and without preventing position bias.}
\label{tab:position_bias}
\end{table}

\subsection{Comparison to Reranking}
To demonstrate the effectiveness of our approach in combining Sent2Sent and Doc2Doc translations, we compare it with two other strategies: 

1) Reranking, which chooses the translation with the higher reference-free COMETKiwi score\footnote{\texttt{wmt22-cometkiwi-da} : \url{https://huggingface.co/Unbabel/wmt22-cometkiwi-da}}~\cite{rei-etal-2022-cometkiwi} for each source sentence~\cite{he-etal-2024-exploring,farinhas-etal-2023-empirical}; 

2) Reranking + Refining, which firstly selects better translation (i.e., Strategy 1) and further refines the selected translation using DocRefine$_\text{doc}$ and DocRefine$_\text{sent}$, similar to \citet{vernikos-etal-2024-dont}.

As shown in Table~\ref{tab:reranking}, our approach outperforms the other two strategies in combining intermediate translations. Furthermore, our approach benefits from the diversity of intermediate translations, achieving the best performance when T1 and T2 originate from Sent2Sent and Doc2Doc \footnote{To generate diverse S2S and D2D translations, we set \texttt{do\_sample} to true, \texttt{temperature} to 0.3 and \texttt{top\_p} to 0.7.}, respectively. This illustrates that our approach effectively integrates the advantages of both translations. For more details, please refer to Appendix \ref{apdx:reranking}.

\begin{table}[]
\centering
\small
\begin{tabular}{lll|cc}
\toprule
\bf T1 & \bf T2 & \bf Strategy & \bf De$\rightarrow$En &\bf En $\rightarrow$ De \\
\hline
S2S & D2D & Rerank  & 86.96 & 84.20  \\
S2S & D2D & Rerank + Refine & 87.74 & 85.56 \\
S2S & D2D & Ours &  \textbf{88.02} & \textbf{86.05} \\
\hline
S2S & S2S & Rerank  & 86.16  & 83.07 \\
S2S & S2S & Rerank + Refine  & 87.63 & 85.58  \\
S2S & S2S & Ours & \bf 87.76 & \bf 86.04 \\
\hline
D2D & D2D & Rerank  & 86.99 & 83.30 \\
D2D & D2D & Rerank + Refine & 87.50 & 85.65 \\
D2D & D2D & Ours & \bf 87.61 & \bf 85.69 \\
\bottomrule
\end{tabular}
\caption{Comparison with \textit{reranking} and \textit{reranking + refining}. T1/T2 refers to intermediate translation 1/2.}
\label{tab:reranking}
\end{table}

\subsection{GPT-based Error Annotating}
\begin{figure}[!t]
\centering
\includegraphics[width=\columnwidth, trim={0cm 0cm 0cm 0cm}]{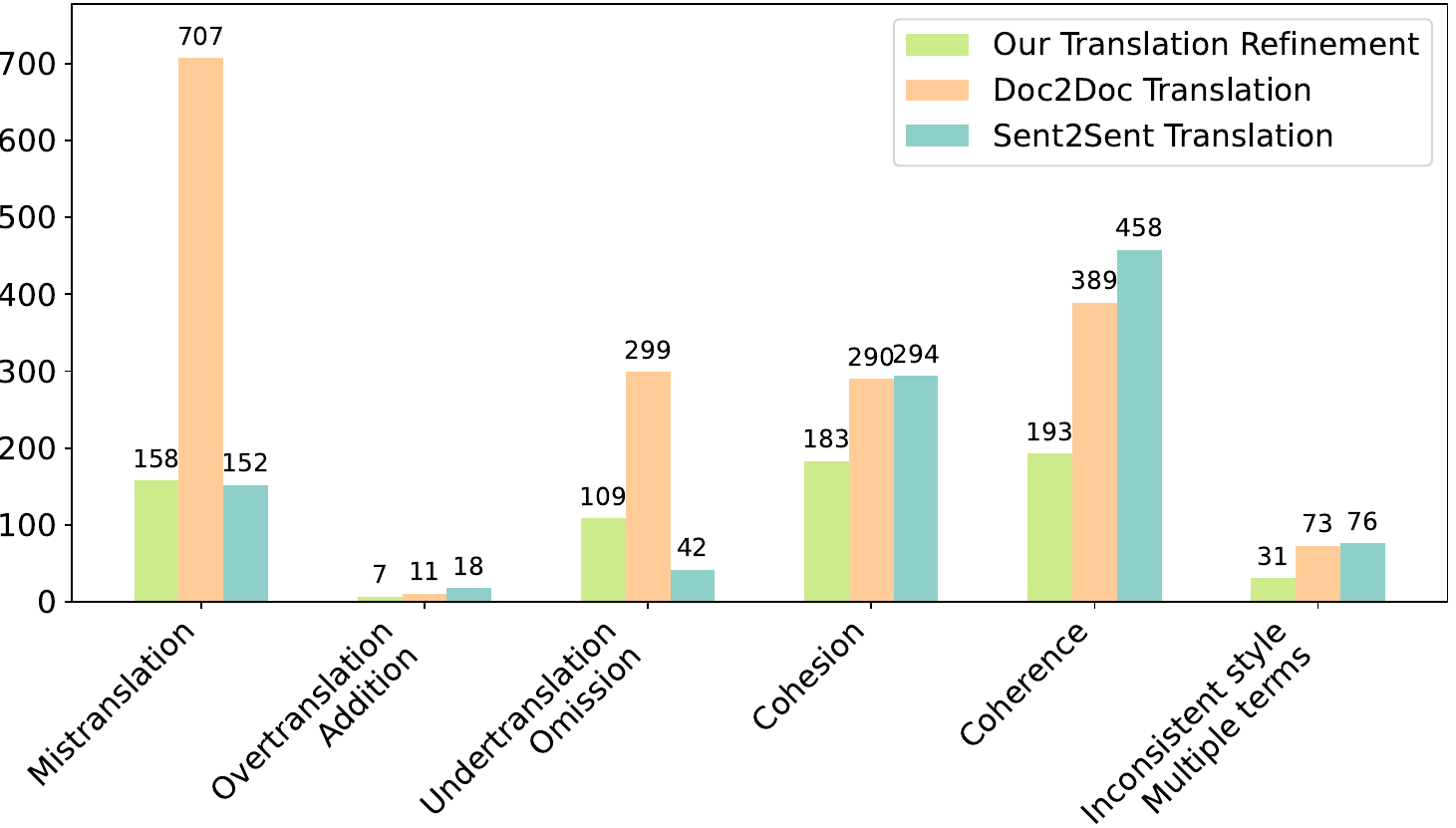}
\caption{Counts of error types on De$\rightarrow$En translation.} 
\label{fig:gpt_bar}
\end{figure}

Following \citet{wu-etal-2024-adapting}, we identify translation errors at both sentence- and document-level. Please refer to Appendix \ref{apdx:mqm} for detailed prompts. Specifically, we use GPT-4o-mini to detect sentence-level issues such as mistranslation, over-translation (including additions), and under-translation (including omissions). Additionally, we address document-level issues related to cohesion, coherence and inconsistent style (including the use of multiple terms for the same concept). Figure~\ref{fig:gpt_bar} shows the results for De$\rightarrow$En translation. It highlights that: 1) our approach addresses all major issues observed in Doc2Doc translation; and 2) it improves most of the issues in Sent2Sent translation, with a trade-off in performance related to under-translation (including omissions). The two highlights suggest that our approach effectively combines the strengths of both Sent2Sent and Doc2Doc translations.

\section{Related Work}
\subsection{LLM-based Translation Refinement}
Current approaches to LLM-based translation refinement can be categorized into two types: prompt engineering and supervised fine-tuning (SFT).

In prompt engineering, \citet{chen-etal-2024-iterative} propose a method where ChatGPT iteratively self-corrects translations. \citet{raunak-etal-2023-leveraging} explore using GPT-4 for automatically post-editing (APE) of neural machine translation (NMT) outputs. \citet{farinhas-etal-2023-empirical} generate multiple hypotheses and experiment with various ensemble methods. \citet{feng-etal-2024-improving} introduce \textit{Translate-Estimate-Refine} framework, leveraging LLMs for self-refinement. \citet{xu-etal-2023-instructscore,xu-etal-2024-llmrefine} prompt LLMs to generate intermediate translations, and then provide self-feedback to optimize the final output. \citet{yang-etal-2023-human} examine human intervention in LLM inference for MT tasks. \citet{chen-etal-2024-dual,chen-etal-2024-ibut} explore LLMs' self-reflective and contextual understanding abilities. \citet{berger-etal-2024-prompting} prompt LLMs to edit translations with human error markings. \citet{chen-etal-2024-refining} apply retrieval-augmented generation (RAG) to enhance translation faithfulness. All of these studies focus on sentence-level refinement. 

In SFT, \citet{ki-carpuat-2024-guiding} train LLMs using source sentences, intermediate translations and error annotations. \citet{alves-etal-2024-tower} fine-tune LLMs for translation-related tasks including APE, and train a model called Tower-Instruct. \citet{feng-etal-2024-ladder} propose hierarchical fine-tuning, grouping instances by refinement difficulty for multi-stage training. While these studies focus on sentence-level refinement, \citet{koneru-etal-2024-contextual} extend refinement by incorporating document-level context. Building on this, our work further extends refinement to the entire document level.

\subsection{LLM-based Document-level Machine Translation}
Current LLM-based document-level machine translation (DMT) approaches can also be categorized into two types: prompt engineering and SFT.

In prompt engineering, \citet{wang-etal-2023-document-level} firstly prompt GPTs for DMT. \citet{karpinska_iyyer_wmt_2023_large} evaluate GPT-3.5 on novel translation tasks. \citet{cui-etal-2024-efficiently} apply RAG to select relevant contextual examples. \citet{wang-etal-2024-delta} and \citet{guo-etal-2025-doc} introduce agents with memory mechanism to capture long-range dependencies to enhance consistency and accuracy. \citet{briakou-etal-2024-translating} frame DMT as a multi-turn process with a step for refinement. \citet{sun-etal-2024-instruction} employ instruction-tuned LLMs and use GPT-4 for document assessment.

On the other hand, SFT approaches enhance LLMs ability for DMT by leveraging tailored training strategies. \citet{li-etal-2024-enhancing} integrate sentence- and document-level instructions. \citet{wu-etal-2024-adapting} introduce a multi-stage fine-tuning approach, first fine-tuning on monolingual documents, then on parallel documents. \citet{stap-etal-2024-fine} fine-tune LLMs on sentence-level instances and evaluate DMT. \citet{lyu-etal-2024-dempt} present a decoding-enhanced, multi-phase prompt tuning method.

\section{Conclusion}
In this paper, we have proposed a novel approach to refine Doc2Doc translation by combining the strengths of both sentence-level and document-level translations. Our approach employs an enhanced fine-tuning with quality awareness to improve the performance of large language models (LLMs). Experimental results across ten document-level translation tasks show substantial improvements in translation quality, coherence, and consistency for a variety of language pairs.

\section*{Limitations}

Our experiments are primarily conducted on a news dataset, which may not fully represent LLMs' performance in other specific domains and other non-English translation directions. Moreover, we train one model for one specific translation direction, leading to huge computational cost. The model may be biased to refining texts of a specific style and may perform worse when refining texts in other styles. Further research may enhance the multilingual performance of LLMs or apply pairwise preference-based optimization tuning.
\bibliography{main}

\appendix
\section{S-COMET Score Distribution}
\label{apdx:distribution}
\begin{figure}[h]
    \centering
    \includegraphics[width=\columnwidth, trim={0cm 0cm 0cm 0cm}]{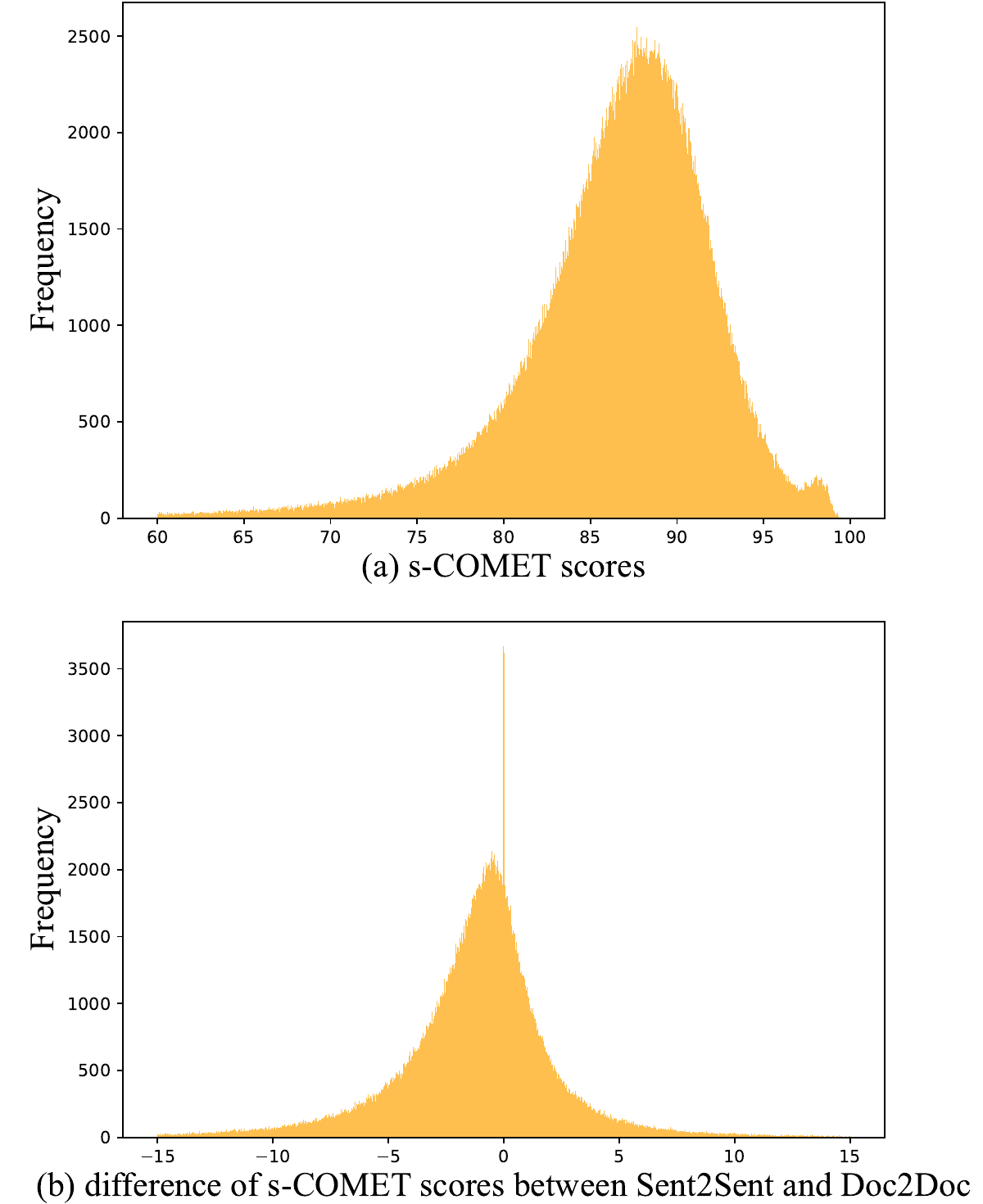}
    \caption{Distribution of s-COMET scores.}
    \label{fig:distribution}
\end{figure}

Figure \ref{fig:distribution} (a) shows the distribution of COMET scores in Chinese (Zh) $\rightarrow$ English (En) dataset produced by LLaMA-3-8B-Instruct. Over 30\% of sentences achieve a s-COMET score above 90.0, while more than 30\% score below 85.0 . 

Figure \ref{fig:distribution} (b) illustrates the distribution of COMET score differences between Sent2Sent and Doc2Doc translations in the same dataset. While some instances exhibit a score difference of zero, the majority follow a normal-like distribution within the range of -15 to 15, with the mean around -1.5 rather than zero.

\section{Data Statistics}
\label{apdx:dataset}

Table~\ref{tab:dataset} shows the detailed statistics of our training, validation and test datasets for the ten translation directions. 
\begin{table}[h]
\centering
\small
\begin{tabular}{c|cc}
\toprule
\textbf{Dataset}& \begin{tabular}[c]{@{}c@{}}\textbf{\#Document}\\ \textbf{Train/Valid/Test}\end{tabular} & \begin{tabular}[c]{@{}c@{}}\textbf{\#Sentence}\\ \textbf{Train/Valid/Test}\end{tabular} \\ 
\hline
De $\leftrightarrow$ En & 8.4K/150/150 & 333K/5.9K/6.0K \\ 
\hline
Fr $\leftrightarrow$ En & 7.9K/150/150 & 310K/5.9K/5.8K \\ 
\hline
Es $\leftrightarrow$ En & 9.7K/150/150 & 378K/5.8K/5.8K \\ 
\hline
Ru $\leftrightarrow$ En & 7.3K/150/150 & 279K/5.7K/5.6K \\ 
\hline
Zh $\leftrightarrow$ En & 8.6K/150/150 & 342K/6.0K/5.9K \\ 
\bottomrule
\end{tabular}
\caption{Statistics of the  datasets.}
\label{tab:dataset}
\end{table}

\section{Details in Splitting Long Documents}
\label{apdx:long_documents}

Similar to~\citet{li-etal-2024-enhancing} and \citet{koneru-etal-2024-contextual}, we split long documents with more than 512 tokens into smaller chunks. Algorithm \ref{alg:splitting documents} denotes the detailed algorithm we use, where $\mathcal M_{\mathcal T}$ denotes the LLM, $N$ denotes the number of the sentences in the document pair, $\mathbf{s}$ denotes the source document, $s_i$ denotes the $i$-th sentence in the document, $L$ denotes the maximum length of the chunk, $\mathbf C$ denotes the list of the chunks in the document, $c$ denotes the chunk, $l_c$ denotes the length of the chunk, respectively. Thus, each document is divided into multiple chunks, each containing no more than 512 tokens while ensuring sentence integrity.

During evaluating document-level metrics, we reassemble the chunks into complete documents.
\renewcommand{\algorithmicrequire}{\textbf{Input:}}
\renewcommand{\algorithmicensure}{\textbf{Output:}}
\begin{algorithm}
\caption{Algorithm for Splitting Documents}\label{alg:splitting documents}
\begin{algorithmic}
\Require $\mathcal M_{\mathcal {T}}$, $N$, $\mathbf s=$ $[s_1,\cdots,s_{N}]$

\Ensure $\mathbf C$

\State $L \gets 512$

\State $\mathbf C \gets []$ 

\State $c \gets []$

\State $l_c \gets 0$

\For{$i \gets 1$ \textbf{to} $N$}
    
    \State $l_i\gets$ the tokenized length of $s_i$ by $\mathcal{M_{\mathcal T}}$
    \If{$l_c+l_i>L$}
    
        \State $\mathbf{C}$.append($c$)
    
        \State $l_c\gets 0$, $c\gets []$
        
    \Comment{Starting a new chunk}
    
        \State $l_c\gets l_i$
        
        \State $c$.append($s_i$)

    \Else
        \State $l_c\gets l_c+l_i$

        \State $c$.append($s_i$)
        
    \EndIf
\EndFor

\State $\mathbf C$.append($c$)

\end{algorithmic}
\end{algorithm}

\section{Discussion on Doc2Doc Translation with Mismatched Source Sentence Boundaries}
\label{apdx:misalignment}
We observe that natural document translations often have mismatched sentence counts between the source and target. Our fine-tuned LLMs handle these cases effectively, as sentence-level alignment is not strictly required during inference. In a small number of cases, this may result in the refined translation having a different number of sentences.

During fine-tuning, only the quality-aware fine-tuning process requires sentence-level alignment between the source and target documents. However, in practical scenarios, this alignment can be relaxed by shifting to segment-level alignment. A segment may consist of one or more sentences, allowing aligned segment pairs to differ in sentence count. For instance, in a parallel document pair $(S, T)$ that is not sentence-aligned, an alignment tool like \texttt{BertAlign}~\cite{liu-zhu-2023-bertalign} can be used to generate sentence-level alignments, which can then be grouped into segment-level alignments.

\section{Fine-Tuning and Inferencing Settings}
\label{apdx:fine_tuning}
During fine-tuning, we adopt QLoRA \cite{dettmers-etal-2024-qlora}, a quantized version of LoRA \cite{hu-etal-2021-lora}. For the hyper-parameters in Eq. \ref{eq:weight}, we set $\lambda$ to 3.75 and $\epsilon$ to 0.7, respectively. we set LoRA rank to 8 and LoRA alpha to 16. We apply LoRA target modules to both the query and the value components. All fine-tuning and inferencing experiments are conducted on 4 NVIDIA V100 GPUs. We use the AdamW optimizer and learning rate scheduler of cosine, with an initial learning rate to 1e-4, warmup ratio of 0.1, batch size of 2, gradient accumulation over 8 steps. In both stages of quality-aware enhanced fine-tuning, we train 1 epoch. During inference, to ensure reproducibility, we set \texttt{do\_sample} to false. Following \citet{alves-etal-2024-tower} and \citet{koneru-etal-2024-contextual}, we set \texttt{num\_beams} to 3. Our implementation is based on LLaMA-Factory Framework\footnote{\url{https://github.com/hiyouga/LLaMA-Factory}} \citep{zheng-etal-2024-llamafactory}.
\section{Effects of Hyper-Parameters}
\label{apdx:hyper_parameter}

\begin{figure}
\centering
\includegraphics[width=\columnwidth, trim={0cm 0cm 0cm 0cm}]{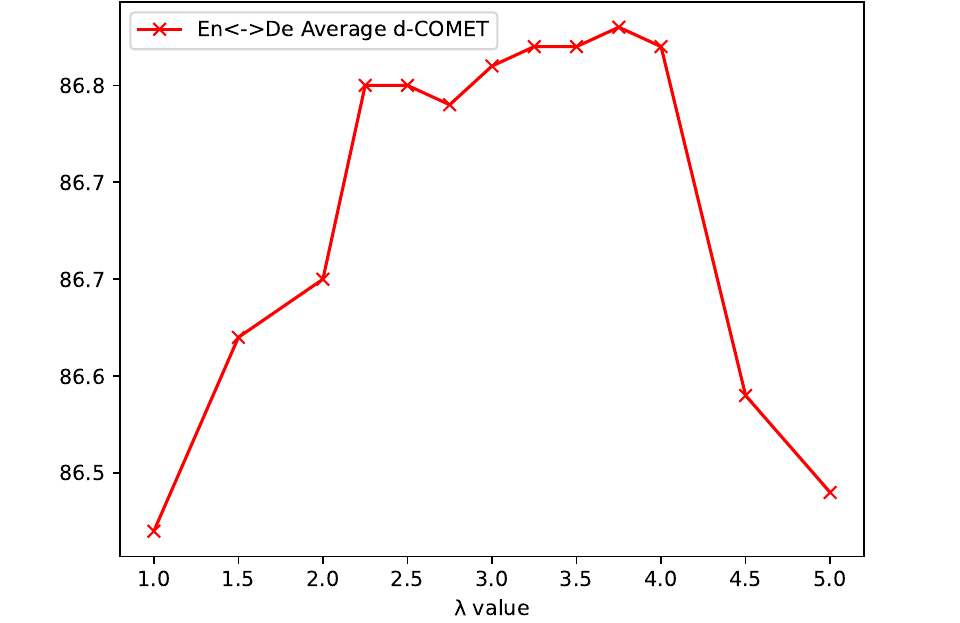}
\caption{Performance of d-COMET scores curve on the En $\leftrightarrow$ De validation sets for $\lambda$ values ranging from 1.0 to 5.0. The optimal performance is achieved when $\lambda = 3.75$. }
\label{fig:hyper_param}
\end{figure} 

We use the combined En $\leftrightarrow$ De validation sets to tune two hyper-parameters: $\lambda$ and $\epsilon$. First, we explore values of $\epsilon$ in the range from 0.5 to 0.9 with a step size of 0.1. Our experiments reveal that $\epsilon$ has a minimal effect on performance, and we ultimately set $\epsilon$ to 0.7.

Next, we search for an optimal value of $\lambda$ within the range of 1.0 to 5.0, using a step size of 0.5. We observe that $\lambda$ values between 2.5 and 4.0 yield better performance than other values. As a result, we narrow the search for $\lambda$ to the range of 2.5 to 4.0 with a finer step size of 0.25. Figure~\ref{fig:hyper_param} illustrates the learning curve for $\lambda$ values between 1.0 and 5.0, showing that $\lambda = 3.75$ achives the best performance.

Based on these findings, we set $\lambda = 3.75$ and $\epsilon = 0.7$ for all experiments.

\section{Comparison to Other Two Weighting Variants}
\label{apdx:weights}
In addition to using Eq.~\ref{eq:weight} to compute the sentence-level weighting, we also compare it with two alternative weighting variants:

\begin{itemize} 
\item Variant 1: Instead of using the maximum \texttt{DA} score, we compute the weight based on $h_i$, which is the first translation in the prompt template (either $y_i$ or $z_i$:): 
\begin{equation} 
\small 
w_i = 1 + \lambda(\texttt{DA}(s_i, h_i, r_i) - \epsilon).
\label{eq:weight_variant1} 
\end{equation} 
\item Variant 2: Rather than assigning a weight to each sentence, we assign a weight to each instance. This instance-level weight is computed as: 
\begin{equation} 
\small 
\begin{split} 
w = 1 + \lambda(\max(&\texttt{avgDA}(s, y, r), \\
&\texttt{avgDA}(s, z, r)) - \epsilon), 
\end{split} 
\label{eq:weight_variant2} 
\end{equation} 
where $\texttt{avgDA}(s, y, r)$ returns the averaged reference-based COMET score. 
\end{itemize}

Table~\ref{tab:comparison_weight} compares the performance. It shows that our weighting method outperforms the other two weighting variants. 

\begin{table}[]
\centering
\small
\begin{tabular}{l|cccc}
\toprule
 & \bf De$\rightarrow$En & \bf En$\rightarrow$De &\bf Zh$\rightarrow$En  &\bf  En$\rightarrow$Zh \\
\hline
Our & \bf 88.14 &  \bf 86.05 & \bf 85.39 &\bf 83.35\\
\hline
Variant 1 & 87.12 & 85.31 & 84.79 & 83.17 \\
Variant 2 & 87.60 & 85.52 & 84.72 & 83.03 \\

\bottomrule
\end{tabular}
\caption{Performance comparison of d-COMET scores when using different equations to calculate weights.}
\label{tab:comparison_weight}
\end{table}
\section{Analyses of Catastrophic Forgetting}
Training models on sentence-level datasets for document-level translation refinement often causes models to generate only the first sentence of the document, leading to catastrophic forgetting. However, our proposed enhanced document-level fine-tuning, incorporating sentence-level quality-aware fine-tuning, preserves the model’s sentence-level translation refinement ability. Specifically, for a given source sentence $s_i$, the model refines two intermediate translations, $y_i$ and $z_i$. As shown in the last row of Table~\ref{tab:catastrophic_forgetting}, our approach maintains strong performance in sentence-level refinement, confirming that catastrophic forgetting is not an issue.

\begin{table*}[h]
\centering
\small
\begin{tabular}{l|ccccc|ccccc|c}
\toprule
\multirow{2}{*}{\bf System} & \multicolumn{5}{c|}{\bf \textit{X}$\rightarrow$En} & \multicolumn{5}{c|}{\bf En$\rightarrow$\textit{X}} & \multirow{2}{*}{\bf Avg.} \\
& \bf De$\rightarrow$ & \bf Es$\rightarrow$ & \bf Ru$\rightarrow$ & \bf Fr$\rightarrow$ & \bf Zh$\rightarrow$ & \bf $\rightarrow$ De & \bf $\rightarrow$ Es & \bf $\rightarrow$ Ru & \bf $\rightarrow$ Fr & \bf $\rightarrow$ Zh\\
\hline
\midrule
Sent2Sent& 85.97 & 86.62 & 81.63 & 84.43 & 82.18 & 82.50 & 85.02 & 80.97 & 82.89 & 76.80 & 82.90 \\
Doc2Doc & 87.05 & 87.21 & 81.07	& 85.40	& 83.60	& 83.35	& 85.36	& 80.18 & 83.14 & 81.89 & 83.83 \\
\hdashline
Ours (document-level) & 88.14 & 88.42 & 82.75 & 86.69 & 86.69 & 86.05 & 86.86 & 83.85 & 84.84 & 83.35 & 85.63 \\ 
Ours (sentence-level) & 87.83 & 88.06 & 82.74 & 86.29 & 82.20 & 86.13 & 84.71 & 83.60 & 84.48 & 82.52 & 84.86 \\
\bottomrule

\end{tabular}
\caption{\label{tab:catastrophic_forgetting} Performance of d-COMET scores when we use LLaMA-3-8B-Instruct to conduct sentence-level refinement with multiple inputs. }
\end{table*}

\section{Analyses of Model-Agnostic}
\label{apdx:model_agnostic}
\begin{table*}[]

\centering
\small
\setlength{\tabcolsep}{5pt}
\begin{tabular}{l|c|c|ccccc|ccccc|c}

\toprule
\bf \multirow{2}{*}{\#} & \multirow{2}{*}{\bf Model 1}& \multirow{2}{*}{\bf Model 2} & \multicolumn{5}{c|}{\bf \textit{X}$\rightarrow$En} & \multicolumn{5}{c|}{\bf En$\rightarrow$\textit{X}} & \multirow{2}{*}{\bf Avg.} \\
&  & & \bf De$\rightarrow$ & \bf Es$\rightarrow$ & \bf Ru$\rightarrow$ & \bf Fr$\rightarrow$ & \bf Zh$\rightarrow$ & \bf $\rightarrow$ De & \bf $\rightarrow$ Es & \bf $\rightarrow$ Ru & \bf $\rightarrow$ Fr & \bf $\rightarrow$ Zh \\
\midrule
1 & LLaMA & Mistral & 87.79 & 88.40 & 82.34 & 86.36 & 85.08 & 86.16 & 86.35 & 83.31 & 84.96 & 82.79 & 85.35 \\
2 & Mistral & LLaMA & 88.01 & 88.38 & 84.25 & 86.61 & 85.28 & 86.79 & 86.95 & 85.95 & 85.50 & 84.12 & 86.18 \\
\bottomrule

\end{tabular}
\caption{\label{tab:refine_others} Performance of d-COMET scores when we use different models in translation and refinement. LLaMA refers to LLaMA-3-8B-Instruct, and Mistral refers to Mistral-Nemo-Instruct.}
\end{table*}

It is not necessary using the same LLM during training and inference. Fine-tuned LLMs can effectively refine translations from other systems, such as GPT-4o-mini and NLLB (Section~\ref{sec:gpt_nllb}). Additionally, LLaMA-3-8B-Instruct can refine Mistral-Nemo-Instruct translations and vice versa. Table~\ref{tab:refine_others} presents the results, where Model 1 generates sentence- and document-level translations, and Model 2 performs refinement.

\section{Translation Refinement Prompts}
\label{apdx:ape_prompts}
\begin{figure}
    \centering
    \includegraphics[width=1\columnwidth]{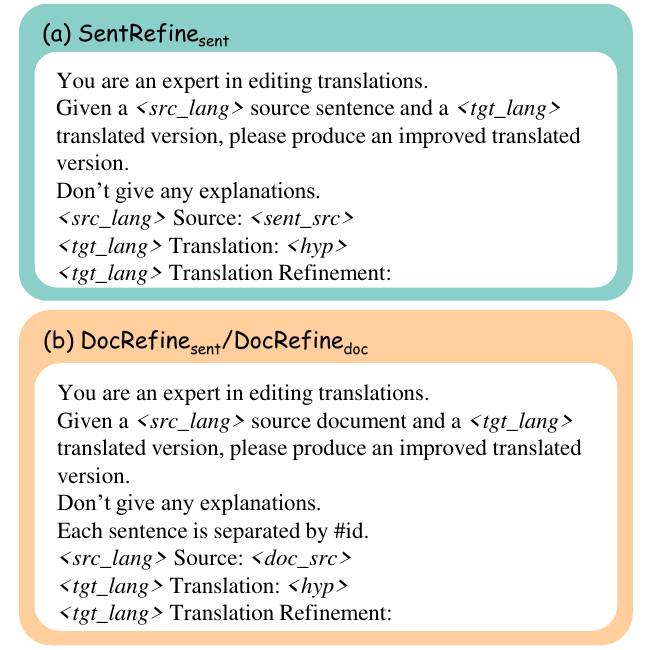}
    \caption{Prompt templates used in our baselines.}
    \label{fig:prompts_apdx}
\end{figure} 
Figure \ref{fig:prompts_apdx} presents the prompt we use for baselines, including SentRefine$_\text{Sent}$, DocRefine$_\text{Sent}$ and DocRefine$_\text{Doc}$. Note that we use the same prompt when we conduct DocRefine$_\text{Sent}$ and DocRefine$_\text{Doc}$.

\section{Experimental Results in s-COMET and d-BLEU}
\label{apdx:other_results}
Table \ref{tab:d-BLEU} shows the detailed d-BLEU scores of our main experiments. Table \ref{tab:s-COMET} shows the detailed s-COMET scores of our main experiments. Table \ref{tab:refine_gpt_scomet} shows the detailed s-COMET scores of our experiments in refining GPT translations.

\begin{table*}[h]
\centering
\small
\begin{tabular}{l|ccccc|ccccc|c}
\toprule
\multirow{2}{*}{\bf System} & \multicolumn{5}{c|}{\bf \textit{X}$\rightarrow$En} & \multicolumn{5}{c|}{\bf En$\rightarrow$\textit{X}} & \multirow{2}{*}{\bf Avg.} \\
& \bf De$\rightarrow$ & \bf Es$\rightarrow$ & \bf Ru$\rightarrow$ & \bf Fr$\rightarrow$ & \bf Zh$\rightarrow$ & \bf $\rightarrow$ De & \bf $\rightarrow$ Es & \bf $\rightarrow$ Ru & \bf $\rightarrow$ Fr & \bf $\rightarrow$ Zh\\
\hline
\multicolumn{12}{c}{LLaMA-3-8B-Instruct} \\ 
\hline
Sent2Sent & 34.73 & 40.81 & 31.16 & 33.30 & 22.35 & 25.07 & 39.33 & 22.25 & 31.85 & 29.12 & 30.99 \\
Sent2Sent$_\text{tuned}$ & \underline{48.26} & 53.44 & 41.58 & 45.09 & 34.02 & \underline{31.93} & 43.92 & 27.24 & 34.29& 36.07 & 39.58\\
Doc2Doc & 37.02 & 43.01 & 32.92 & 34.52 & 26.33 & 25.68 & 40.04 & 23.09 & 30.32 & 33.41 & 32.63 \\
Doc2Doc$_\text{tuned}$ & 47.04 & 53.50& 42.80 & 43.35 & 35.95 & 30.11 & 44.59 & 27.37 & 34.96 & 38.65 & 39.83 \\
\hdashline
SentRefine$_\text{sent}$  &46.11 &52.54 &42.20& 43.58& 32.88&30.22&44.84& 27.38&\underline{35.05}& 38.07& 39.29\\
DocRefine$_\text{sent}$ & 45.16 & 53.77 & 44.33 & \underline{45.44} & 35.92 & 30.02 & 43.93 & 26.68 & 34.90 & 37.79 & 39.79 \\
DocRefine$_\text{doc}$ & 46.16 & 53.90 & 44.32 & 45.07 & 36.14 & 29.50 & 44.65 & 28.34 & 34.73 & 37.65 & 40.05 \\ \hdashline
Ours & \textbf{48.51} & \textbf{54.70} & \textbf{45.59} & \textbf{45.57} & \textbf{37.66} & \textbf{32.23} & \textbf{45.78} & \textbf{28.74} & \textbf{35.26} & \textbf{38.96} & \textbf{41.30} \\ 
~~ - \footnotesize{QA Fine-tuning}& 47.86 & \underline{54.07} & \underline{44.81} & 45.02 & \underline{37.07} & 31.47 & \underline{44.87} & \underline{28.43} & 34.42 & \underline{38.77} & \underline{40.68} \\ 
\hline
\multicolumn{12}{c}{Mistral-Nemo-Instruct}\\ 
\hline
Sent2Sent & 38.18 & 43.20 & 34.45 & 35.87 & 27.51 & 29.02 & 41.88 & 25.44 & 33.17 & 34.37 & 34.31 \\
Sent2Sent$_\text{tuned}$ & 40.62 & 45.67 & 39.29 & 38.93 & 31.90 & 30.00 & 42.77 & 27.15 &33.73 & 35.07 & 36.51\\
Doc2Doc & 40.92 & 45.20 & 37.51 & 37.98 &29.74& 29.70 & 42.10 & 27.88 & 34.10 &37.09& 36.22 \\
Doc2Doc$_\text{tuned}$ & 49.17 &55.10 & 43.35 & 46.01 & \underline{38.25} & 31.65 & 45.75 & 22.15& 37.10 & 42.24 & 41.08 \\
\hdashline
SentRefine$_\text{sent}$ &46.11&52.54& 47.90 &45.25&32.65&30.22&44.84&30.40&36.05& 35.10&40.11\\
DocRefine$_\text{sent}$      & 48.75 & 55.56 & 46.45 & 46.49 &36.76& 34.13 & 46.12 & 31.13 & 37.45 &41.44& 42.43 \\
DocRefine$_\text{doc}$       & 49.77 & \underline{55.70} & 46.29 & \underline{46.52} & 37.09 & 33.82 & 46.33 & 31.02 & 37.29 &\underline{42.68}& 42.65 \\ \hdashline
Ours & \textbf{51.17} & \textbf{56.20} & \textbf{48.58} & \textbf{47.97} & \textbf{41.00} & \textbf{35.44} & \textbf{47.01} & \textbf{32.79} & \textbf{38.43} & \textbf{43.13} & \textbf{44.17} \\ 
~~ - \footnotesize{QA Fine-tuning} & \underline{50.43} & 55.37 & \underline{47.97} & 45.92 & 37.89 & \underline{35.28} & 
\underline{46.64} & \underline{31.62} & \underline{37.87} & 42.41 & 
\underline{43.14} \\
\hline
\end{tabular}
\caption{Performance in document-level (d-BLEU) score.}
\label{tab:d-BLEU}
\end{table*}

\begin{table*}[]
\centering
\small
\begin{tabular}{l|ccccc|ccccc|c}
\toprule
\multirow{2}{*}{\bf System} & \multicolumn{5}{c|}{\bf \textit{X}$\rightarrow$En} & \multicolumn{5}{c|}{\bf En$\rightarrow$\textit{X}} & \multirow{2}{*}{\bf Avg.} \\
& \bf De$\rightarrow$ & \bf Es$\rightarrow$ & \bf Ru$\rightarrow$ & \bf Fr$\rightarrow$ & \bf Zh$\rightarrow$ & \bf $\rightarrow$ De & \bf $\rightarrow$ Es & \bf $\rightarrow$ Ru & \bf $\rightarrow$ Fr & \bf $\rightarrow$ Zh\\
\hline
\multicolumn{12}{c}{LLaMA-3-8B-Instruct} \\ 
\hline
Sent2Sent & 87.71 & 88.32 & 83.74 & 86.63 & 84.60 & 84.47 & 86.82 & 83.23 & 84.55 & 79.76 & 84.98\\
Sent2Sent$_\text{tuned}$ &88.93 & 88.91 & 86.38 & 88.33 & 86.27 & 86.28 & 87.12 & 86.25 & \underline{86.43} & 86.49 & 87.14 \\
Doc2Doc  &88.62&88.76& 84.47 &87.36&85.84&83.87&87.07&82.61&84.79&83.85&85.72\\
Doc2Doc$_\text{tuned}$ & 89.35 & 89.91 & 80.51 & 88.29 &  86.38 & 87.20 & 88.20 & 83.76 & 86.26 & 85.51 & 86.54\\
\hdashline
SentRefine$_\text{sent}$ & 89.12&89.65& \textbf{85.29} &88.08&86.53&87.10&  88.17 &\textbf{87.16}&86.38& \underline{86.70} &\underline{87.42}\\
DocRefine$_\text{sent}$  &88.96&89.08& 83.09 &\underline{88.45}& 87.19 &87.18& 88.17 &83.21&86.08& 86.42& 86.78\\
DocRefine$_\text{doc}$  &89.22& 89.51& 84.45 &88.24& \underline{87.25}&86.86&  88.34 &86.12&86.39&\underline{86.70}&87.31 \\ \hdashline
Ours &\textbf{89.63} & \textbf{89.95} & \underline{84.58} & \textbf{88.58}& \textbf{87.26}&\textbf{87.76}&\textbf{88.61}& \underline{86.34} &\textbf{86.50}&\textbf{86.88}&\textbf{87.61} \\
~~ - \footnotesize{QA Fine-tuning} & \underline{89.41} & \underline{89.88} & 84.44 & 88.43 & 87.19 & \underline{87.43} & \underline{88.37} & 85.63 & 86.14 & 86.69 & 87.36 \\ 
\hline
\multicolumn{12}{c}{Mistral-Nemo-Instruct}\\ 
\hline
Sent2Sent & 88.52 & 88.40 & 84.24 & 87.00 & 86.18 & 86.64 & 87.32 & 86.25 & 85.52 & 85.41 & 86.54 \\
Sent2Sent$_\text{tuned}$ & 88.49 & 88.55 & 85.03 & 87.78 & 86.42 & 87.24 & 87.22 & 87.17 & 86.52 & 85.85 & 87.03\\
Doc2Doc & 89.15 & 89.29 & 85.16 & 87.90 & 86.81 & 86.56 & 87.30 & 86.66 & 85.65 & 85.74 & 87.02 \\
Doc2Doc$_\text{tuned}$ & 89.70 & \underline{90.20} & 85.01& 88.61 & 87.70 & 85.91 & 88.66 & 85.19 & 86.99 & 87.56 & 87.53 \\
\hdashline
SentRefine$_\text{sent}$ & 89.33 & 89.80 & 85.51 & 88.24 & 86.71 & 88.04 & 88.30 & \underline{87.89} & 86.77 & 86.71 & 87.73 \\
DocRefine$_\text{sent}$  & 89.63 & 90.03 & 84.21 & 88.02 & 87.64 & 88.24 & 88.68 & 86.93 & 86.90 & 87.55 & \underline{87.78} \\
DocRefine$_\text{doc}$ & \underline{89.74} & 90.06 & 83.50 & 88.21 & 87.49 & 88.21 & 88.69 & 86.33 & 86.85 & 87.44 & 87.65  \\ \hdashline
Ours & \textbf{89.94} & \textbf{90.45} & \textbf{86.10} & \underline{88.51} & \textbf{87.96} & \textbf{88.53} & 
\textbf{89.02} & \textbf{88.31} & \textbf{87.16} & \textbf{88.04} & \textbf{88.40} \\
~~ - \footnotesize{QA Fine-tuning} & 89.90& 90.12 & \underline{85.82} & \textbf{88.65} & \underline{87.87} & \underline{88.49} & \underline{88.87} & 87.81 & \underline{87.07} & \underline{87.71} & 88.23\\
\hline
\end{tabular}
\caption{Performance in sentence-level COMET (s-COMET) score.}
\label{tab:s-COMET}
\end{table*}

\begin{table*}[ht]
\centering
\small
\setlength{\tabcolsep}{5pt}
\begin{tabular}{l|l|ccccc|ccccc|c}
\toprule
\bf \multirow{2}{*}{\#} & \multirow{2}{*}{\bf System} & \multicolumn{5}{c|}{\bf \textit{X}$\rightarrow$En} & \multicolumn{5}{c|}{\bf En$\rightarrow$\textit{X}} & \multirow{2}{*}{\bf Avg.} \\
 & & \bf De$\rightarrow$ & \bf Es$\rightarrow$ & \bf Ru$\rightarrow$ & \bf Fr$\rightarrow$ & \bf Zh$\rightarrow$ & \bf $\rightarrow$ De & \bf $\rightarrow$ Es & \bf $\rightarrow$ Ru & \bf $\rightarrow$ Fr & \bf $\rightarrow$ Zh\\
\hline
\multicolumn{13}{c}{GPT Translation \& Refining GPT Translation}\\
\hline
1 & GPT Sent2Sent  & 88.39 & 88.51 & 83.76 & 87.05 & 86.34& 87.43 & 87.63 & 87.55 &  86.35 & 87.09 & 87.01 \\
2 & GPT Doc2Doc & 88.12 & 89.10 & 85.24 & 87.02 & 86.96 & 87.98 & 88.41 & 87.88 & 86.83 & 87.70 & 87.52 \\
\hdashline
3 & GPT SentRefine$_\text{sent}$ & 88.51 & 88.56 & 84.42 & 87.63 & 86.60  & 87.72 & 88.28 & 87.16 & 86.72 & 87.44 & 87.30 \\
4 & GPT DocRefine$_\text{sent}$ & 88.65 & 88.69 & 84.82 & 87.61 & 86.55 & 88.41  & 88.78 & 88.45 & 87.10 & 87.24 & 87.63 \\
5 & GPT DocRefine$_\text{doc}$ & 88.64 & 88.90 & 84.83 & 87.70 & 86.65 & 88.38 & 88.71 & \underline{88.47} &\underline{87.16} & 87.42 & 87.69\\ 
6 & GPT DocRefine$_\text{doc+sent}$ & \underline{88.99} & 89.25 & 85.09 & 87.79 & 86.98 & 88.28 & 88.59 & 88.41 & 87.03 & \underline{87.79} & \underline{87.82} \\
\hdashline
7 & L-DocRefine$_\text{doc+sent}$ & 88.78 & 88.98 & 84.28 & \underline{87.81} & 86.73 & 88.16 & \bf 89.11 & 86.45 & 86.95 & 87.32 & 87.46 \\
8 & M-DocRefine$_\text{doc+sent}$ &\bf 90.02 & 89.05 &\bf 86.29 & \bf 87.92 & \underline{86.99} & \bf 88.67 & \underline{89.08} & \bf 88.57 & \bf 87.32 & \bf 87.95 & \bf 88.19 \\
\hline
\multicolumn{13}{c}{NLLB Translation \& Refining NLLB Translation}\\
\hline
9& NLLB Sent2Sent& 88.45& 89.16 & 84.89 & 87.66 & 85.65 & 86.74 & 88.30& 87.76 & 86.57& 77.82 & 86.29\\
\hdashline
10& L-DocRefine$_\text{doc+sent}$&88.76& \underline{89.27}& 83.92& 87.74& \bf 87.35& 88.14& 88.44& 85.79&86.99& 86.80& 87.32 \\
11& M-DocRefine$_\text{doc+sent}$& 88.98&\bf 89.50& \underline{85.33}& \bf 87.92& 86.92& \underline{88.60}& 88.08& 88.31& \bf 87.32& 86.96& 87.79\\
\hline

\bottomrule

\end{tabular}
\caption{\label{tab:refine_gpt_scomet}Performance in s-COMET when refining translations from GPT-4o-mini. For the GPT-based refinement systems, we use the same prompt templates as those used in our approach, but without fine-tuning. L-* and M-* denote our fine-tuned LLaMA-3-8B-Instrcut and Mistral-Nemo-Instruct, respectively.}
\end{table*}

\section{Comparison of Our Approach with Reranking Variant}

\label{apdx:reranking}
Since our approach uses two intermediate translations, we compare it to a reranking variant that selects the better sentence-level translation from our two baselines, ensuring a fair comparison. Specifically, we calculate the percentage of sentences, based on the reference-based COMET score, where our approach either outperforms, underperforms, or ties\footnote{If the difference in their COMET scores is 0.1 or smaller, the two translations are considered a tie.} with the reranking variant.

\begin{figure}
    \centering
    \includegraphics[width=\columnwidth, trim={0cm 0cm 0cm 0cm}]{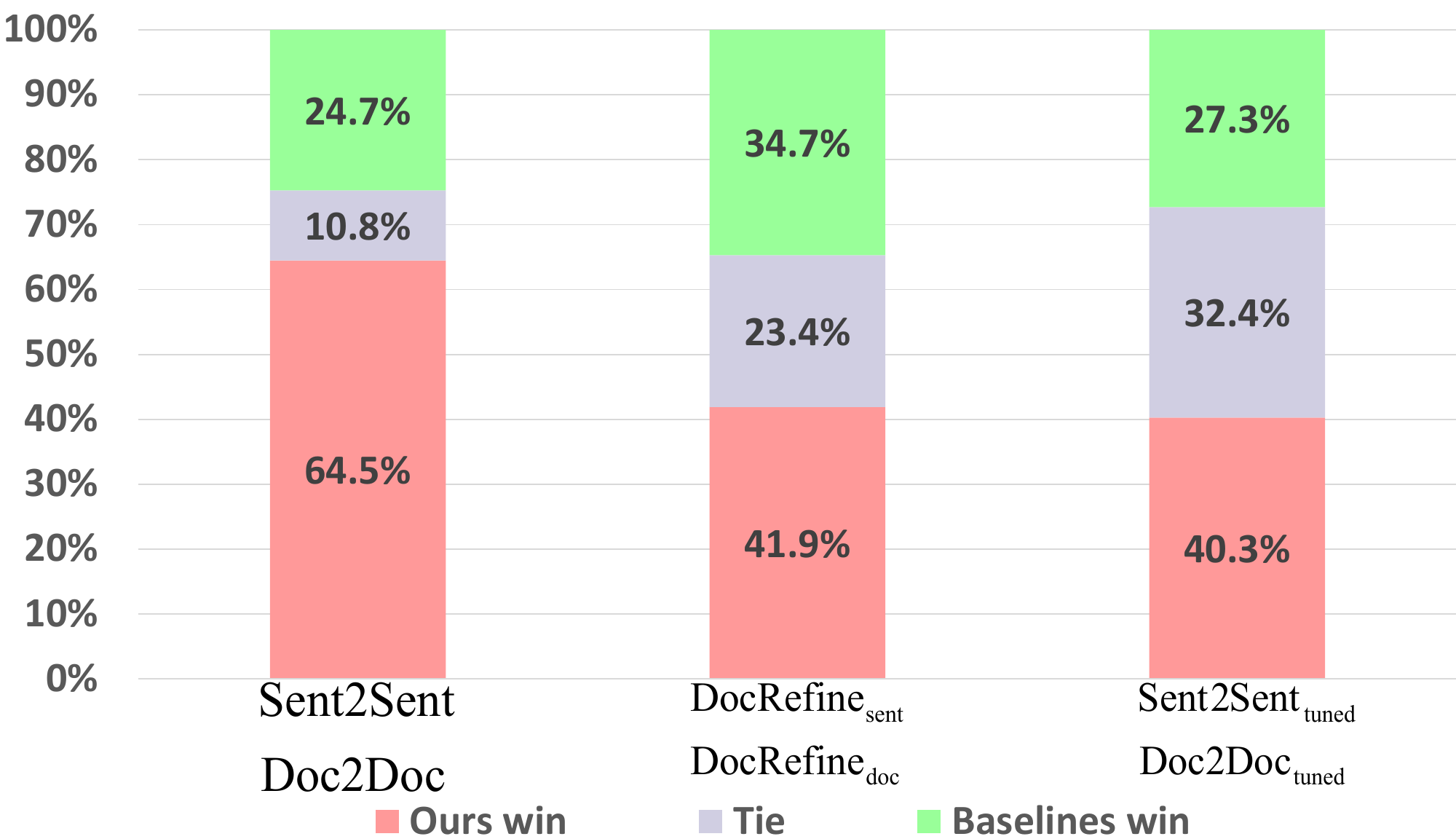}
    \caption{Comparison of our approach with the reranking variant.}
    \label{fig:bar}
\end{figure}

Figure~\ref{fig:bar} presents the comparison results for De $\leftrightarrow$ En translation. It demonstrates that our approach outperforms the reranking variant by winning more sentences, even when the latter reranks several different two baselines.

\section{Prompt for Analysing Translation Errors}

\label{apdx:mqm}
We present the prompt used for analysing translation errors in Table \ref{tab:error_analyses_prompt}. "Mistranslation", "Overtranslation", "Undertranslation", "Addition" and "Omission" are sentence-level translation error types, while "Cohesion", "Coherence", "Inconsistent style" and "Multiple terms in translation" are document-level translation error types. 

\begin{table*}[]
\begin{tabular}{|p{\textwidth}|}
\hline
{[}Source{]}:\\ \textit{<src\_doc>}\\ 
{[}Reference{]}:\\ \textit{<ref\_doc>}\\
{[}Hypothesis{]}:\\ \textit{<hyp\_doc>} \\
\\
{[}Error Types{]}: \\ - Mistranslation: Error occurring when the target content does not accurately represent the source. \\ - Overtranslation: Error occurring in the target content that is inappropriately more specific than the source.\\ - Undertranslation: Error occurring in the target content that is inappropriately less specific than the source.\\ - Addition: Error occurring in the target content that includes content not present in the source. \\ - Omission: Error where content present in the source is missing in the target.  \\ - Cohesion: Portions of the text needed to connect it into an understandable whole (e.g., reference, substitution, ellipsis, conjunction, and lexical cohesion) missing or incorrect.\\ - Coherence: Text lacking a clear semantic relationship between its parts, i.e., the different parts don't hang together, don't follow the discourse conventions of the target language, or don't "make sense."\\ - Inconsistent style: Style that varies inconsistently throughout the text, e.g., One part of a text is written in a clear, "terse" style, while other sections are written in a more wordy style.\\ - Multiple terms in translation: Error where source content terminology is correct, but target content terms are not used consistently.\\ \\ Considering the provided context, please identify the errors of the translation from the source to the target in the current sentence based on a subset of Multidimensional Quality Metrics (MQM) error typology. \\ You should pay extra attention to the error types related to the relationship between the current sentence and its context, such as "Unclear reference", "Cohesion", "Coherence", "Inconsistent style", and "Multiple terms in translation". \\ For each sentence in machine translation, please give the error types and brief explanation for errors.The returned format is as follows:\\ Sentence \#id : \\ Error types: ...\\ Explanation for errors: ... \\

\hline
\end{tabular}
\caption{\label{tab:error_analyses_prompt} Prompt used for analyzing translation errors. }
\end{table*}

\end{document}